\documentclass{article} 
\usepackage{iclr2024_conference,times}

\usepackage{amsmath,amsfonts,bm}

\def\eqref#1{equation~\ref{#1}}

\def\1{\bm{1}}

\DeclareMathAlphabet{\mathsfit}{\encodingdefault}{\sfdefault}{m}{sl}
\SetMathAlphabet{\mathsfit}{bold}{\encodingdefault}{\sfdefault}{bx}{n}

\usepackage{hyperref}
\usepackage{url}
\usepackage{times}
\usepackage{epsfig}
\usepackage{graphicx}
\usepackage{amsmath}
\usepackage{bbding}
\usepackage{pifont}
\usepackage{wasysym}
\usepackage{amssymb}
\usepackage{bm}
\usepackage{epsfig}
\usepackage{graphicx}
\usepackage{amsmath}
\usepackage{float}
\usepackage{enumitem}
\usepackage{xcolor}
\usepackage{wrapfig}
\usepackage{caption}
\usepackage{makecell}
\usepackage[ruled]{algorithm2e}

\newcommand{\circled}[1]{\textcircled{\raisebox{-.9pt}{#1}}}

\title{DragonDiffusion: Enabling \underline{Drag}-style Manipulation \underline{on} \underline{Diffusion} Models}

\author{Antiquus S.~Hippocampus, Natalia Cerebro \& Amelie P. Amygdale \thanks{ Use footnote for providing further information
about author (webpage, alternative address)---\emph{not} for acknowledging
funding agencies.  Funding acknowledgements go at the end of the paper.} \\
Department of Computer Science\\
Cranberry-Lemon University\\
Pittsburgh, PA 15213, USA \\
\texttt{\{hippo,brain,jen\}@cs.cranberry-lemon.edu} \\
\And
Ji Q. Ren \& Yevgeny LeNet \\
Department of Computational Neuroscience \\
University of the Witwatersrand \\
Joburg, South Africa \\
\texttt{\{robot,net\}@wits.ac.za} \\
\AND
Coauthor \\
Affiliation \\
Address \\
\texttt{email}
}

\begin{document}

\maketitle

\begin{abstract} 
Despite the ability of existing large-scale text-to-image (T2I) diffusion models to generate high-quality images from detailed textual descriptions, they often lack the ability to precisely edit the generated or real images. In this paper, we propose a novel image editing method, \textbf{DragonDiffusion}, enabling \textbf{Drag}-style manipulation \textbf{on} \textbf{Diffusion} models. Specifically, we treat image editing as the change of feature correspondence in a pre-trained diffusion model. By leveraging feature correspondence, we develop energy functions that align with the editing target, transforming image editing operations into gradient guidance. Based on this guidance approach, we also construct multi-scale guidance that considers both semantic and geometric alignment. Furthermore, we incorporate a visual cross-attention strategy based on a memory bank design to ensure consistency between the edited result and original image. Benefiting from these efficient designs, all content editing and consistency operations come from the feature correspondence without extra model fine-tuning or additional modules. Extensive experiments demonstrate that our method has promising performance on various image editing tasks, including editing within a single image (\textit{e.g.}, object moving, resizing, and content dragging) and across images (\textit{e.g.}, appearance replacing and object pasting).

\end{abstract}

\section{Introduction}
Thanks to the large-scale training data and huge computing power, generative models have developed rapidly, especially large-scale text-to-image (T2I) diffusion models~\cite{t2i1,ldm,glid,dall-e2}, which aims to generate images conditioned on a given text/prompt. However, this generative capability is usually diverse, and it is challenging to design suitable prompts to generate images consistent with what the user has in mind, let alone fine-grained image editing based on the text condition.

In the community of image editing, previous methods are usually designed based on GANs~\cite{abdal2019image2stylegan, abdal2020image2stylegan++, alaluf2022hyperstyle} due to the compact and editable latent space, \textit{e.g.}, the $\mathcal{W}$ space in StyleGAN~\cite{karras2019style}. Recently, DragGAN~\cite{draggan} proposes a point-to-point dragging scheme, which can achieve refined content dragging. However, it is limited by the capacity and generalization of GANs. Compared to GANs, diffusion model~\cite{diff} has higher stability and superior generation quality. 
Due to the lack of a concise and editable latent space, numerous diffusion-based image editing methods~\cite{p2p, edit1, edit2} are built based on T2I diffusion models via correspondence between text and image features. 
Recently, self-guidance~\cite{selfG} proposes a differentiable approach that employs cross-attention maps between text and image to locate and calculate the size of objects within images. Then, gradient guidance is utilized to edit these properties. However, the correspondence between text and image features is weak, heavily relying on the design of prompts. Moreover, in complex or multi-object scenarios, text struggles to build accurate correspondence with a specific object. In this paper, we aim to investigate whether the diffusion model can achieve drag-style image editing, which is a fine-grained and generalized editing ability not limited to point dragging.

In the large-scale T2I diffusion model, besides the correspondence between text features and intermediate image features, there is also a strong correspondence across image features.
This characteristic is studied in DIFT~\cite{dift}, which demonstrates that this correspondence is high-level, enabling point-to-point correspondence of relevant image content. Therefore, we are intrigued by the possibility of utilizing this strong correspondence across image features to achieve image editing. 
In this paper, we regard image editing as the change of feature correspondence and convert it into gradient guidance via energy functions~\cite{clf} in score-based diffusion~\cite{score}. Additionally, the content consistency between editing results and original images is also ensured by feature correspondence in a visual cross-attention design.
Here, we notice that there is a concurrent work, DragDiffusion~\cite{dragdiff}, studying this issue. It uses LORA~\cite{lora} to maintain consistency with the original image and optimizes the latent in a specific diffusion step
to perform point dragging. Unlike DragDiffusion, our image editing is achieved by energy functions and a visual cross-attention design,
without the need for extra model fine-tuning or new blocks. In addition, we can complete various
drag-style image editing tasks beyond the point dragging. 

In summary, the contributions of this paper are as follows: 

\begin{itemize}
    \item We achieve drag-style image editing via gradient guidance produced by image feature correspondence in the pre-trained diffusion model. In this design, we also study the roles of the features in different layers and develop multi-scale guidance that considers both semantic and geometric correspondence.
    
    \item We design a memory bank, further utilizing the image feature correspondence to maintain the consistency between editing results and original images. In conjunction with gradient guidance, our method allows a direct transfer of T2I generation ability in diffusion models to image editing tasks without the need for extra model fine-tuning or new blocks.

    \item Extensive experiments demonstrate that our method has promising performance in various image editing tasks, including editing within a single image (\textit{e.g.}, object moving, resizing, and content dragging) or across images (\textit{e.g.}, appearance replacing and object pasting).
\end{itemize}

\section{Related Work}

\subsection{Diffusion Models}
Recently, diffusion models~\cite{diff} have achieved great success in the community of image synthesis. It is designed based on thermodynamics~\cite{phy1,phy2}, including a diffusion process and a reverse process. In the diffusion process, a natural image $\mathbf{x}_0$ is converted to a Gaussian distribution $\mathbf{x}_T$ by adding random Gaussian noise with $T$ iterations. 
The reverse process is to recover $\mathbf{x}_0$ from $\mathbf{x}_T$ by several denoising steps. Therefore, the diffusion model is to train a denoiser, conditioned on the current noisy image $\mathbf{x}_{t}$ and time step $t$:
\begin{equation}
\label{loss}
    \mathbb{E}_{\mathbf{x}_{0},t, \boldsymbol{\epsilon}_t \sim \mathcal{N}(0,1)}\left[ ||\boldsymbol{\epsilon}_t-\epsilon_{\boldsymbol{\theta}}(\mathbf{x}_{t},t)||_2^2\right],
\end{equation}
where $\epsilon_{\boldsymbol{\theta}}$ is the function of the denoiser. Recently, some text-conditioned diffusion models (\textit{e.g.}, GLID~\cite{glid} and StableDiffusion(SD)~\cite{ldm}) are proposed, which mostly inject text condition into the denoiser through a cross-attention strategy. From the continuous perspective~\cite{score}, diffusion models can be viewed as a score function (\textit{i.e.}, $\epsilon_{\boldsymbol{\theta}}(\mathbf{x}_{t},t) \approx \nabla_{\mathbf{x}_t} \log q(\mathbf{x}_t)$) that samples from the corresponding distribution~\cite{score1} according to Langevin dynamics~\cite{phy1,phy2}.

\subsection{Energy Function in Diffusion Model} 
From the continuous perspective of score-based diffusion, the external condition $\mathbf{y}$ can be combined by a conditional score function, \textit{i.e.}, $\nabla_{\mathbf{x}_t} \log q(\mathbf{x}_t | \mathbf{y})$, to sample from a more enriched distribution.
The conditional score function can be further decomposed as:
\begin{equation}
\label{score}
\small
    \nabla_{\mathbf{x}_t} \log q(\mathbf{x}_t | \mathbf{y}) =  \log\left(\frac{q(\mathbf{x}_t | \mathbf{y})q(\mathbf{x}_t)}{q(\mathbf{y})}\right) \propto \nabla_{\mathbf{x}_t} \log q(\mathbf{x}_t)+\nabla_{\mathbf{x}_t} \log q(\mathbf{y}|\mathbf{x}_t),
\end{equation}
where the first term is the unconditional denoiser, and the second term refers to the conditional gradient produced by an energy function $\mathcal{E}(\mathbf{x}_t; t, \mathbf{y}) = q(\mathbf{x}_t | \mathbf{y})$.
$\mathcal{E}$ can be selected based on the generation target, such as a classifier~\cite{clf} to specify the category of generation results. Energy function has been used in various controllable generation tasks, \textit{e.g.}, sketch-guided generation \cite{nvidia}, mask-guided generation~\cite{mask-guided}, universal guidance \cite{freedom, universal}, and image editing~\cite{selfG}. 
These methods, inspire us to transform editing operations into conditional gradients, achieving fine-grained image editing.

\subsection{Image Editing}
In image editing, numerous previous methods~\cite{abdal2019image2stylegan, abdal2020image2stylegan++, alaluf2022hyperstyle} invert images into the latent space of StyleGAN~\cite{karras2019style} and then edit the image by manipulating latent vectors.  
Motivated by the success of diffusion model~\cite{diff}, various diffusion-based image editing methods~\cite{avrahami2022blended, p2p,kawar2023imagic,meng2021sdedit,brooks2023instructpix2pix} are proposed. Most of them use text as the editing control. For example, \cite{kawar2023imagic, valevski2023unitune, kwon2022diffusion} perform model fine-tuning on a single image and then generate the editing result by target text. 
Prompt2Prompt~\cite{p2p} achieves specific object editing by exchanging text-image attention maps. SDEdit~\cite{meng2021sdedit} performs image editing by adding noise to the original image and then denoising under new text conditions. InstructPix2Pix~\cite{brooks2023instructpix2pix} retrain the diffusion model with text as the editing instruction. Recently, Self-guidance~\cite{selfG} transforms image editing operations into gradients through the correspondence between text and image features. However, text-guided image editing is coarse. Recently, DragGAN~\cite{draggan} proposes a point-to-point dragging scheme, which can achieve fine-grained dragging editing. Nevertheless, its editing quality and generalization are limited by GANs. 
How to utilize the high-quality and diverse generation ability of diffusion models for fine-grained image editing is still an open challenge.

\section{Method}
\subsection{Preliminary: How to Construct Energy Function in Diffusion}
\label{eng}
Modeling an energy function $\mathcal{E}(\mathbf{x}_t; t, \mathbf{y})$ to produce the conditional gradient $\nabla_{\mathbf{x}_t} \log q(\mathbf{y}|\mathbf{x}_t)$ in Eq.~\ref{score}, remains an open question. $\mathcal{E}$ measures the distance between $\mathbf{x}_t$ and the condition $\mathbf{y}$. Some methods \cite{clf,nvidia,egsde} train a time-dependent distance measuring function, \textit{e.g.}, a classifier~\cite{clf} to predict the probability that $\mathbf{x}_t$ belongs to category $\mathbf{y}$. However, the training cost and annotation difficulty are intractable in our image editing task. 
Some tuning-free methods~\cite{freedom, universal} propose using the clean image $\mathbf{x}_{0|t}$ predicted at each time step $t$ to replace $\mathbf{x}_{t}$ for distance measuring, \textit{i.e.}, $\mathcal{E}(\mathbf{x}_t; t, \mathbf{y}) \approx \mathcal{D}(\mathbf{x}_{0|t}; t, \mathbf{y})$. Nevertheless, there is a bias between $\mathbf{x}_{0|t}$ and $\mathbf{x}_{0}$, and there is hardly a suitable $\mathcal{D}$ for distance measuring in image editing tasks.
Hence, the primary issue is whether we can circumvent the training requirement and construct an energy function to measure the distance between $\mathbf{x}_{t}$ and the editing target. Recent work \cite{dift} has shown that the feature correspondence in the diffusion UNet-denoiser $\epsilon_{\mathbf{\theta}}$ is high-level, enabling point-to-point correspondence measuring. Inspired by this characteristic, we propose reusing $\epsilon_{\mathbf{\theta}}$ as a tuning-free energy function to transform image editing operations into the change of feature correspondence. 

\subsection{Overview}
\begin{figure*}[t]
\centering
\small 
\begin{minipage}[t]{\linewidth}
\centering
\includegraphics[width=1\columnwidth]{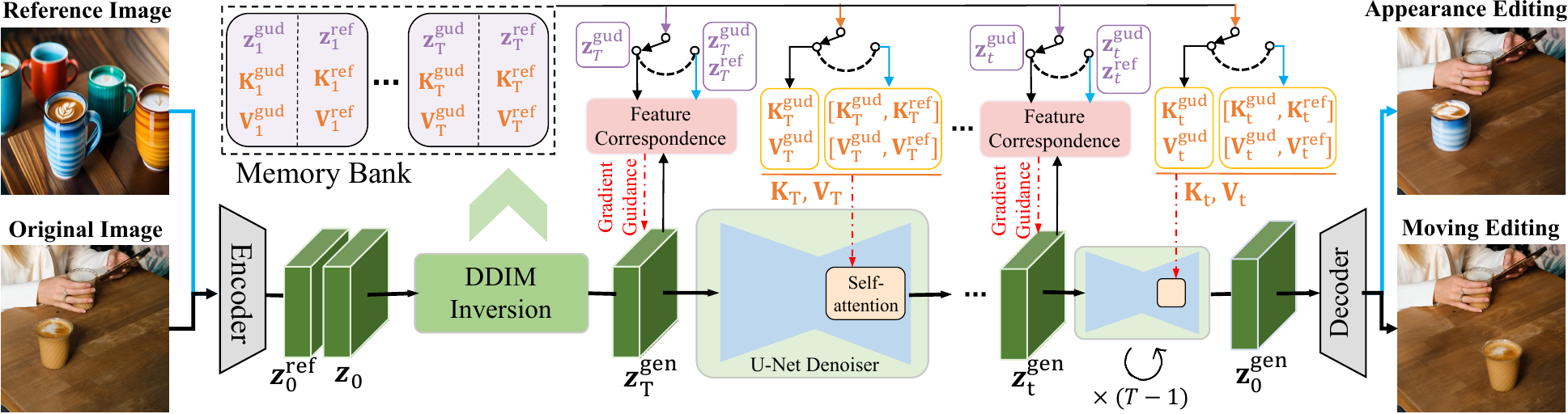}
\end{minipage}
\centering

\caption{Overview of our DragonDiffusion, which consists of two parts: (1)DDIM inversion~\cite{ddim_inv} to build a memory bank; (2)inference with guidance from the memory bank. Our method is built on the pre-trained SD~\cite{ldm} without extra training or modules. 
}
\label{network} 
\end{figure*}

The editing objective of our DragonDiffusion
involves two issues: changing the content to be edited and preserving unedited content. For example, if a user wants to move the cup in an image, the generated result only needs to change the position of the cup, while the appearance of the cup and other unedited content should not change. 
An overview of our method is presented in Fig.~\ref{network}, which is built on the pre-trained SD~\cite{ldm} to support image editing with and without reference images. First, we use DDIM inversion~\cite{ddim_inv} to transform the
original image into $\mathbf{z}_T$. If the reference image $\mathbf{z}_0^{ref}$ exists, it will also be involved in the inversion. In this process, we store some intermediate features and latent at each time step to build a memory bank, which is used to provide guidance for subsequent image editing. 
In generation, 
we transform the information stored in the memory bank into content editing and consistency guidance through two paths, \textit{i.e.}, visual cross-attention and gradient guidance. Both of these paths are built based on feature correspondence in the pre-trained SD.
Therefore, our image editing pipeline is efficiently built without extra model fine-tuning or new blocks.

\subsection{DDIM Inversion with Memory Bank}
\label{mem_bank}
In our image editing process, the starting point $\mathbf{z}_{T}$, produced by DDIM inversion~\cite{ddim_inv}, can provide a good generation prior to maintain consistency with the original image. 
However, relying solely on the final step of this approximate inversion can hardly provide accurate generation guidance. Therefore, we fully utilize the information in DDIM inversion by building a memory bank 
to store the latent $\mathbf{z}_t^{gud}$ at each inversion step $t$, as well as corresponding keys $\mathbf{K}_t^{gud}$ and values $\mathbf{V}_t^{gud}$ in the self-attention module of the decoder within the UNet denoiser. 
Note that in some cross-image editing tasks (\textit{e.g.}, appearance replacing, object pasting), reference images are required. In these tasks, the memory bank needs to be doubled to store the information of the reference images. Here, we utilize $\mathbf{z}_t^{ref}$, $\mathbf{K}_t^{ref}$, and $\mathbf{V}_t^{ref}$ to represent them.
The information stored in the memory bank will provide more accurate guidance for the subsequent image editing process. 

\begin{wrapfigure}{r}{0.42\textwidth}
    \centering
    \vspace{-40pt}
    \includegraphics[width=0.42\textwidth]{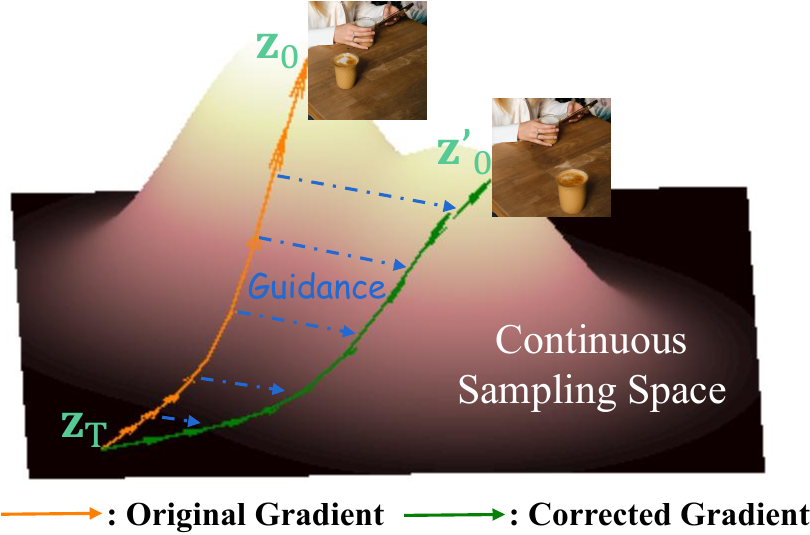}
    \vspace{-20pt}
    \caption{\small{Illustration of continuous sampling space in score-based diffusion. Bright colors indicate areas where target data is densely distributed. The orange and green paths respectively refer to the diffusion paths without and with external gradient guidance.}}
    \vspace{-30pt}
    \label{fig_gradient} 
\end{wrapfigure}
\subsection{Gradient-guidance-based Editing Design}
Inspired by classifier guidance~\cite{clf}, we build energy functions to transform image editing operations into gradient guidance in diffusion sampling. 
An intuitive illustration is presented in Fig.~\ref{fig_gradient}, showing a continuous sampling space of the score-based diffusion~\cite{score}. 
The sampling starting point $\mathbf{z}_T$, obtained from DDIM inversion, will approximately return to
the original point only according to the gradient/score predicted by the denoiser. After incorporating the gradient guidance generated by the energy function that matches the editing target, the additional guidance gradient will change the path to reach a sampling result that meets the editing target.

\subsubsection{Energy Function via Feature Correspondence}
In our DragonDiffusion, energy functions are designed to provide gradient guidance for image editing, mainly including content editing and consistency terms.
Specifically, at the $t$-th time step, we reuse the UNet denoiser $\epsilon_{\boldsymbol{\theta}}$ to extract intermediate features $\mathbf{F}_{t}^{gen}$ from the latent $\mathbf{z}_{t}^{gen}$ at the current time step. The same operation is used to extract guided features $\mathbf{F}_{t}^{gud}$ from 
$\mathbf{z}_{t}^{gud}$ in memory bank. 
Following DIFT~\cite{dift}, $\mathbf{F}_{t}^{gen}$ and $\mathbf{F}_{t}^{gud}$ come from intermediate features in the UNet decoder. 
The image editing operation is represented by two binary masks (\textit{i.e.}, $\mathbf{m}^{gud}$ and $\mathbf{m}^{gen}$) to locate the original content position and target dragging position, respectively.
Therefore, the energy function is built by constraining the correspondence between these two regions in $\mathbf{F}_{t}^{gud}$ and $\mathbf{F}_{t}^{gen}$. Here, we utilize cosine distance $\cos(\cdot)\in [-1,1]$ to measure the similarity and normalize it to $[0,1]$:
\begin{equation}
\label{sim_local}
    \mathcal{S}_{local}(\mathbf{F}_{t}^{gen}, \mathbf{m}^{gen}, \mathbf{F}_{t}^{gud}, \mathbf{m}^{gud} )=0.5\cdot \cos\left(\mathbf{F}_{t}^{gen}[\mathbf{m}^{gen}],\ \text{sg}(\mathbf{F}_{t}^{gud}[\mathbf{m}^{gud}])\right)+0.5,
\end{equation}
where $\text{sg}(\cdot)$ is the gradient clipping operation. 
Eq.~\ref{sim_local} is mainly used for dense constraints on the spatial location of content. In addition, a global appearance similarity is defined as:
\begin{equation}
\label{sim_global}
    \mathcal{S}_{global}(\mathbf{F}_{t}^{gen}, \mathbf{m}^{gen},  \mathbf{F}_{t}^{gud}, \mathbf{m}^{gud}) = 0.5\cdot \cos\left(\frac{\sum\mathbf{F}_{t}^{gen}[\mathbf{m}^{gen}]}{\sum \mathbf{m}^{gen}},\ \text{sg}(\frac{\sum \mathbf{F}_{t}^{gud}[\mathbf{m}^{gud}]}{\sum \mathbf{m}^{gud}})\right)+0.5,
\end{equation}
which utilizes the mean of the features in a region as a global appearance representation. 
When we want to have fine control over the spatial position of an object or a rough global control over its appearance, we only need to constrain the similarity in Eq.~\ref{sim_local} and Eq.~\ref{sim_global} to be as large as possible. Therefore, the energy function to produce editing guidance is defined as:  
\begin{equation}
\label{loss_edit}
    \mathcal{E}_{edit} = \frac{1}{\alpha+\beta \cdot \mathcal{S}(\mathbf{F}_{t}^{gen}, \mathbf{m}^{gen}, \mathbf{F}_{t}^{gud}, \mathbf{m}^{gud})},\ \ \ \ \mathcal{S}\in \{\mathcal{S}_{local},\ \mathcal{S}_{global}\},
\end{equation}
where $\alpha$ and $\beta$ are two hyper-parameters, which are set as $1$ and $4$, respectively. In addition to editing, we hope the unedited content remains consistent with the original image. We use a mask $\mathbf{m}^{share}$ to locate areas without editing. The similarity between the editing result and the original image in $\mathbf{m}^{share}$ can also be calculated by the cosine similarity as $\mathcal{S}_{local}(\mathbf{F}_{t}^{gen}, \mathbf{m}^{share}, \mathbf{F}_{t}^{gud}, \mathbf{m}^{share})$. Therefore, the energy function to produce content consistency guidance is defined as:
\begin{equation}
\label{loss_con}
    \mathcal{E}_{content} = \frac{1}{\alpha+\beta \cdot \mathcal{S}_{local}(\mathbf{F}_{t}^{gen}, \mathbf{m}^{share}, \mathbf{F}_{t}^{gud}, \mathbf{m}^{share})}.
\end{equation}

In addition to $\mathcal{E}_{edit}$ and $\mathcal{E}_{content}$, an optional guidance term $\mathcal{E}_{opt}$ may need to be added in some tasks to achieve the editing goal. Finally, the base energy function 
is defined as:
\begin{equation}
\label{eq_loss}
    \mathcal{E} = w_{e}\cdot \mathcal{E}_{edit} + w_{c}\cdot \mathcal{E}_{content} + w_{o}\cdot \mathcal{E}_{opt},
\end{equation}
where $w_{e}$, $w_{c}$, and $w_{o}$ are hyper-parameters to balance these guidance terms.
They vary slightly in different editing tasks but are fixed within the same task. Finally, regarding $[\mathbf{m}^{gen}, \mathbf{m}^{share}]$ as condition, the conditional score function in Eq.~\ref{score} can be written as:
\begin{align}
\label{score_our}
\begin{split}
    \nabla_{\mathbf{z}_t^{gen}} \log q(\mathbf{z}_t^{gen} | \mathbf{y}) \propto 
    \nabla_{\mathbf{z}_t^{gen}} \log q(\mathbf{z}_t^{gen})+\nabla_{\mathbf{z}_t^{gen}} \log q(\mathbf{y}|\mathbf{z}_t^{gen}),\ \ \ \mathbf{y}=[\mathbf{m}^{gen}, \mathbf{m}^{share}].
\end{split}
\end{align}
The conditional gradient 
$\nabla_{\mathbf{z}_t^{gen}} \log q(\mathbf{y}|\mathbf{z}_t^{gen})$ can be computed by $\nabla_{\mathbf{z}_t^{gen}} \mathcal{E}$, which will also multiplies by a learning rate $\eta$. In experiments, we find that the gradient guidance in later diffusion generation steps hinders the generation of textures. Therefore, we only add gradient guidance in the first $n$ steps of diffusion generation. Experientially, we set $n=30$ in $50$ sampling steps.

\begin{figure*}[t]
\centering
\small 
\begin{minipage}[t]{\linewidth}
\centering
\includegraphics[width=1\columnwidth]{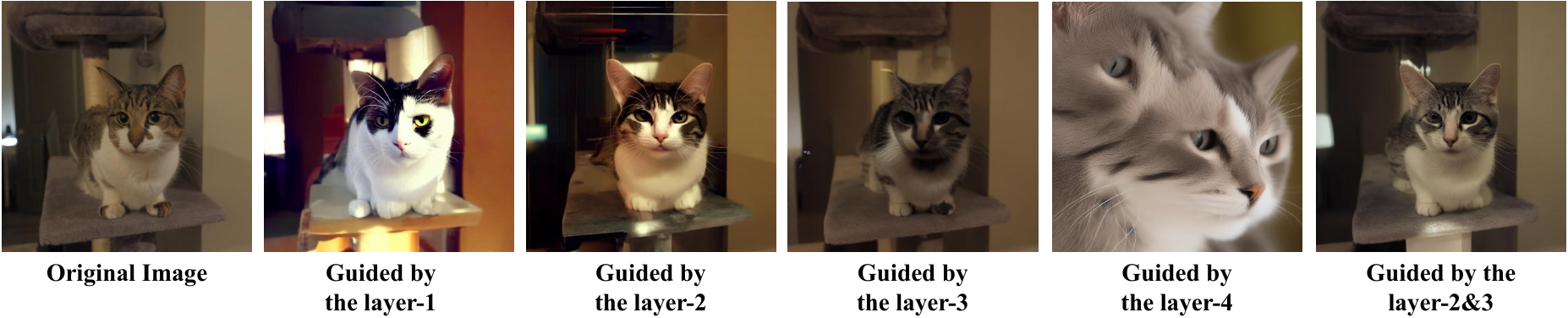}
\end{minipage}
\centering
\vspace{-15pt}
\caption{Illustration of using features from different layers as guidance to restore the original image. $\mathbf{z}_T$ is randomly initialized. The generation is solely guided by content consistency guidance in Eq.~\ref{loss_con}. 
}
\label{layer} 
\end{figure*}

\subsubsection{Multi-scale Feature Correspondance}
The decoder of the UNet denoiser contains four blocks of different scales. DIFT~\cite{dift} finds that the second layer contains more semantic information, while the third layer contains more geometric information. We also studied the role of features from different layers in image editing tasks, as shown in Fig.~\ref{layer}. In the experiment, we set $\mathbf{z}_T$ as random Gaussian noise and set $\mathbf{m}^{gen}$, $\mathbf{m}^{gud}$ as zeros matrixes. $\mathbf{m}^{share}$ is set as a ones matrix. In this way, generation relies solely on content consistency guidance (\textit{i.e.}, Eq.~\ref{loss_con}) to restore image content. We can find that the guidance from the first layer is too high-level to reconstruct the original image accurately. The guidance from the fourth layer has weak feature correspondence, resulting in significant differences between the reconstructed and original images. The features from the second and third layers are more suitable to produce guidance signals, and each has its own specialty. Concretely, the features in the second layer contain more semantic information and can reconstruct images that are semantically similar to the original image but with some differences in content details. The features in the third layer tend to express low-level characteristics, but they cannot provide effective supervision for high-level texture, resulting in blurry results. In our design, we combine these two levels (\textit{i.e.}, high and low) of guidance by proposing a multi-scale supervision approach. Specifically, we compute gradient guidance on the second and third layers. The reconstructed results in Fig.~\ref{layer} also demonstrate that this combination can balance the generation of low-level and high-level visual characteristics. 

\begin{wrapfigure}{r}{0.52\textwidth}
    \centering
    \vspace{-15pt}
    \includegraphics[width=0.52\textwidth]{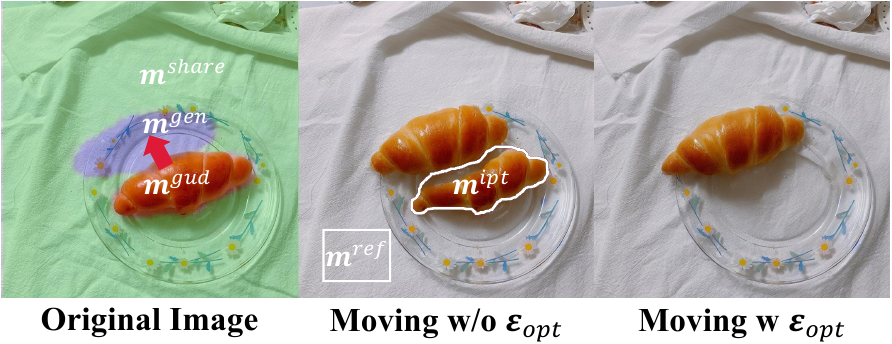}
    \vspace{-15pt}
    \caption{Visualization of the effectiveness of inpainting guidance ($\mathcal{E}_{opt}$) in the object moving task, presenting that $\mathcal{E}_{opt}$ can guide the inpainting of the area where the object is initially located.}
    \vspace{-10pt}
    \label{fig_loss} 
\end{wrapfigure}
\subsubsection{Implementation Details for Each Application}
\textbf{Object moving.} In the task of object moving, $\mathbf{m}^{gen}$ and $\mathbf{m}^{gud}$ locate the same object in different spatial positions. $\mathbf{m}^{share}$ is the complement ($\text{C}_{\text{u}}$) of the union ($\cup$) of $\mathbf{m}^{gen}$ and $\mathbf{m}^{gud}$, \textit{i.e.}, $\mathbf{m}^{share}=\text{C}_{\text{u}}(\mathbf{m}^{gen}\cup \mathbf{m}^{gud})$. However, solely using the content editing and consistency guidance in Eq.~\ref{loss_edit} and Eq.~\ref{loss_con} can lead to some issues, as shown in the second image of Fig.~\ref{fig_loss}. Concretely, although the bread is moved according to the editing signal, some of the bread content is still preserved in its original position in the generated result. 
This is because the energy function does not constrain the area where the moved object was initially located, causing inpainting to easily restore the original object. 
To rectify this issue, we use the optional energy term (\textit{i.e.}, $\mathcal{E}_{opt}$ in Eq.~\ref{eq_loss}) to constrain the inpainting content to be dissimilar to the moved object and similar to a predefined reference region.
Here, we use $\mathbf{m}^{ref}$ to locate the reference region and define
$\mathbf{m}^{ipt} =\{p|p\in \mathbf{m}^{gud}\ and\ p\notin \mathbf{m}^{gen}\} $ to locate the inpainting region. Finally, $\mathcal{E}_{opt}$ in this task is defined as:
\begin{align}
\small
\label{loss_inpaint}
\begin{split}
    \mathcal{E}_{opt}=\frac{w_{i}}{\alpha + \beta\cdot \mathcal{S}_{global}(\mathbf{F}_{t}^{gen}, \mathbf{m}^{ipt}, \mathbf{F}_{t}^{gud}, \mathbf{m}^{ref})}+ \mathcal{S}_{local}(\mathbf{F}_{t}^{gen}, \mathbf{m}^{ipt}, \mathbf{F}_{t}^{gud}, \mathbf{m}^{ipt}),
\end{split}
\end{align}
where $w_{i}$ is a weight parameter, set as $2.5$ in our implementation. 
The third image in Fig.~\ref{fig_loss} shows that this design can effectively achieve the editing goal without impeachable artifact.

\noindent \textbf{Object resizing.} The score function in this task is the same as the object moving, except that a scale factor $\gamma>0$ is added during feature extraction. 
Specifically, we use interpolation to transform $\mathbf{m}^{gud}$ and $\mathbf{F}^{gud}_t$ to the target size, and then extract $\mathbf{F}^{gud}_t[\mathbf{m}^{gud}]$ as the feature of the resized object. To locate the target object with the same size in $\mathbf{F}^{gen}_t$, we resize $\mathbf{m}^{gen}$ with the same scale factor $\gamma$. Then we extract a new $\mathbf{m}^{gen}$ of the original size from the center of the resized $\mathbf{m}^{gen}$. Note that if $\gamma<1$, we use $0$ to pad the vacant area. 

\begin{wrapfigure}{r}{0.41\textwidth}
    \centering
    \vspace{-10pt}
    \includegraphics[width=0.41\textwidth]{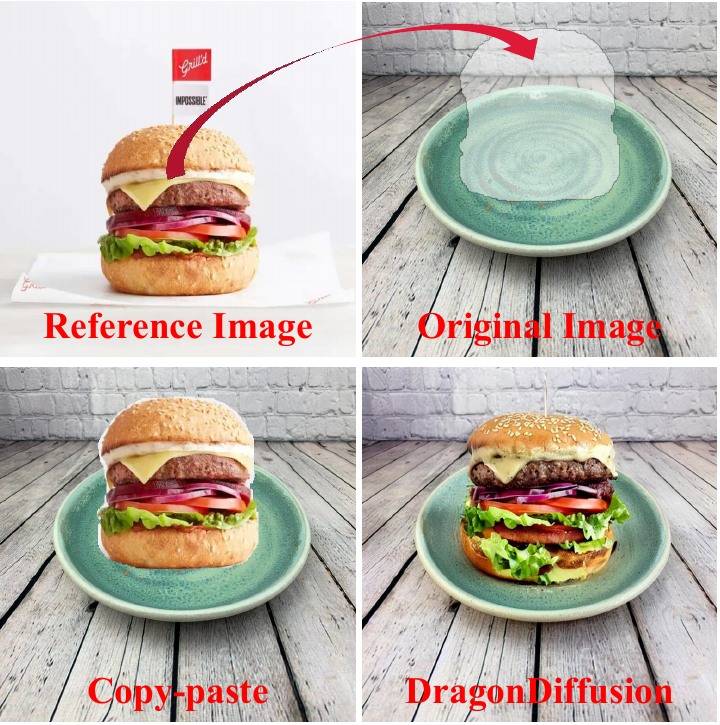}
    \vspace{-20pt}
    \caption{Visual comparison between our DragonDiffusion and direct \textit{copy-paste} in cross-image object pasting.}
    \vspace{-15pt}
    \label{fig_paste} 
\end{wrapfigure}
\noindent \textbf{Appearance replacing.} This task aims to replace the appearance between objects of the same category across images. Therefore, the capacity of the memory bank needs to be doubled to store extra information from the image containing the reference appearance, \textit{i.e.}, $\mathbf{z}_{t}^{ref}$, $\mathbf{K}_{t}^{ref}$, and $\mathbf{V}_{t}^{ref}$. $\mathbf{m}^{gen}$ and $\mathbf{m}^{gud}$ respectively locate the editing object in the original image and the reference object in the reference image. $\mathbf{m}^{share}$ is set as the complement of $\mathbf{m}^{gen}$, \textit{i.e.}, $\text{C}_{\text{u}}(\mathbf{m}^{gen})$. To constrain appearance, we choose $\mathcal{S}_{global}(\mathbf{F}_{t}^{gen}, \mathbf{m}^{gen}, \mathbf{F}_{t}^{ref}, \mathbf{m}^{gud})$ in Eq.~\ref{loss_edit}. In this task, there is no need for $\mathcal{E}_{opt}$.

\noindent \textbf{Object pasting.} 
Object pasting aims to paste an object from an image onto any position in another image. Although it can be completed by simple \textit{copy-paste}, it often results in inconsistencies between the paste area and other areas due to differences in light and perspective, as shown in Fig.~\ref{fig_paste}. As can be seen, the result obtained by \textit{copy-paste} exists discontinuities, while the result generated by our DragonDiffusion can achieve a more harmonized integration of the scene and the pasted object. In implementation, similar to the appearance replacing, the memory bank needs to store information of the reference image, which contains the target object. $\mathbf{m}^{gen}$ and $\mathbf{m}^{gud}$ respectively mark the position of the object in the edited image and reference image. $\mathbf{m}^{share}$ is set as $\text{C}_{\text{u}}(\mathbf{m}^{gen})$. 

\noindent \textbf{Point dragging.} In this task, we want to drag image content via several points, like DragGAN~\cite{draggan}. In this case, $\mathbf{m}^{gen}$ and $\mathbf{m}^{gud}$ locate neighboring areas centered around the destination and starting points. Here, we extract a $3\times3$ rectangular patch centered around each point as the neighboring area.
Unlike the previous tasks, $\mathbf{m}^{share}$ here is manually defined. 

\subsection{Visual Cross-attention}
As mentioned previously, two strategies are used to ensure the consistency between the editing result and the original image: (1) DDIM inversion to initialize $\mathbf{z}_T$; (2) content consistency guidance in Eq.~\ref{loss_con}. However, it is still challenging to maintain high consistency. Inspired by the consistency preserving in some video and image editing works~\cite{tune,masactrl,video-zero}, we design a visual cross-attention guidance. Instead of generating guidance information through an independent inference branch, we reuse the intermediate features of the inversion process stored in the memory bank. Specifically, similar to the injection of text conditions in SD~\cite{ldm}, we replace the key and value in the self-attention module of the UNet decoder with the corresponding key and value collected by the memory bank in DDIM inversion. 
Note that in the appearance replacing and object pasting tasks, the memory bank stores two sets of keys and values from the original image ($\mathbf{K}^{gud}_t, \mathbf{V}^{gud}_t$) and the reference image ($\mathbf{K}^{ref}_t, \mathbf{V}^{ref}_t$). In this case, we concatenate the two sets of keys and values in the length dimension.
The visual cross-attention at each time step is defined as:
\begin{align}
\begin{split}
\small
\left \{
\begin{array}{ll}
    \mathbf{Q}_t = \mathbf{Q}^{gen}_t;\ \mathbf{K}_t = \mathbf{K}^{gud}_t\ \text{or}\ (\mathbf{K}^{gud}_t \circled{\textbf{c}} \mathbf{K}^{ref}_t);\ \mathbf{V}_t = \mathbf{V}^{gud}_t\ \text{or}\ (\mathbf{V}^{gud}_t \circled{\textbf{c}} \mathbf{V}^{ref}_t) \\
    \text{Att}(\mathbf{Q}_t, \mathbf{K}_t, \mathbf{V}_t) = \text{softmax}(\frac{\mathbf{Q}_t\mathbf{K}_t^T}{\sqrt{d}}) \mathbf{V}_t,
\end{array}
\right.
\end{split}
\end{align}
where $\circled{\textbf{c}}$ refers to the concatenation operation.

\begin{figure*}[t]
\centering
\small 
\begin{minipage}[t]{\linewidth}
\centering
\includegraphics[width=1\columnwidth]{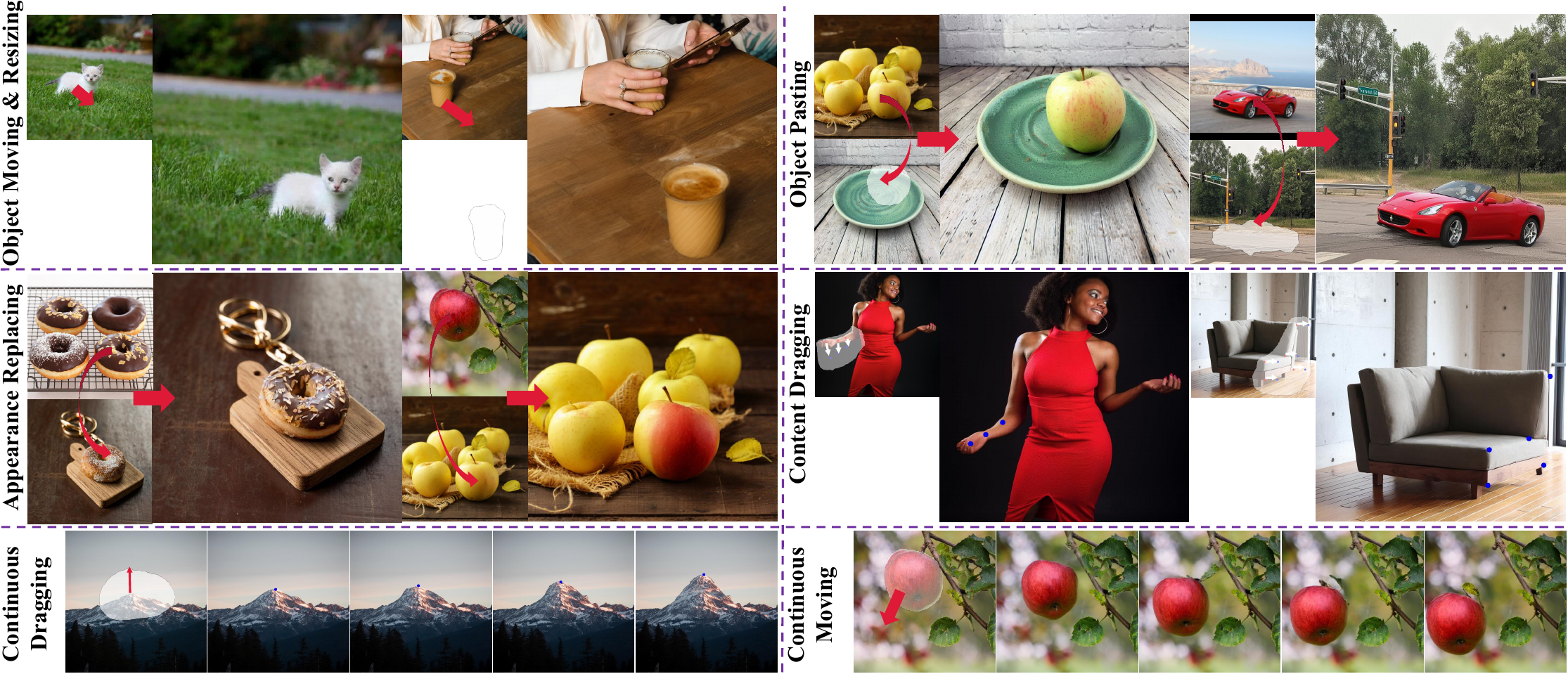}
\end{minipage}
\centering
\vspace{-15pt}
\caption{Visualization of our method on different image editing applications, including object moving, object resizing, object appearance replacing, object pasting, and content dragging.}
\label{fig_application} 
\end{figure*}

\section{Experiments}
In experiments, we use StableDiffusion-V1.5~\cite{ldm} as the base model. The inference adopts DDIM sampling with 50 steps, and we set the classifier-free guidance scale as 5. 
\subsection{Applications}
In this paper, our proposed DragonDiffusion can perform various image editing tasks without specific training. These applications include object moving, resizing, appearance replacing, object pasting, and content dragging. In Fig.~\ref{fig_application}, we present our editing performance on each application. The object moving and resizing in the first block show that our method can naturally move and resize objects in images with good content consistency. The moved objects can blend well with the surrounding content. The second block shows that our method can paste an object from an image into another image and slightly adjust the appearance to blend in with new scenarios. In the third block, we present the performance of object appearance replacing. It shows that our method can replace the appearance with that of a same-category object from a reference image while preserving the original outline. The fourth block shows that our method can drag the content within the image with several points. The dragging results are reasonable with the editing direction, while the result remains consistent with the original image. The last row in Fig.~\ref{fig_application} presents the stable performance in continuous editing. More results are presented in the \textbf{appendix}.

\begin{table}[t]
\caption{Quantitative evaluation on face manipulation with 68 and 17 points. The accuracy is calculated by Euclidean distance between edited points and target points. The initial distance (\textit{i.e.}, \textit{57.19} and \textit{36.36}) is the upper bound, without editing.
FID~\cite{fid} is utilized to quantize the editing quality of different methods. The time complexity is computed on the ‘1 point’ dragging.}
\vspace{-5pt}
\small
\centering
\setlength{\tabcolsep}{4.2pt}
\begin{tabular}{c | c c c c c c}
\hline
 & \makecell[c]{Preparing\\complexity$\downarrow$} & \makecell[c]{Inference\\complexity$\downarrow$} & \makecell[c]{Unaligned\\face} & \makecell[c]{17 Points$\downarrow$\\From 57.19} & \makecell[c]{68 Points$\downarrow$\\ From 36.36} & \makecell[c]{FID$\downarrow$\\17/68 points} \\
\hline
UserControllableLT & \textbf{1.2}s & \textbf{0.05}s & \XSolidBrush & 32.32 & 24.15 & 51.20/50.32 \\
DragGAN & 52.40s & \underline{6.71s} & \XSolidBrush & \textbf{15.96} & \textbf{10.60} & 39.27/39.50\\
DragDiffusion & 48.25s & 19.71s & \Checkmark & 22.95 & 17.32 & \underline{38.06}/\underline{36.55}\\
DragonDiffusion(ours) & \underline{3.62s} & 15.93s & \Checkmark & \underline{18.51} & \underline{13.94} & \textbf{35.75}/\textbf{34.58}\\
\hline
\end{tabular}
\label{tb_cp_face}
\vspace{-5pt}
\end{table}

\begin{figure*}[t]
\centering
\small 
\begin{minipage}[t]{\linewidth}
\centering
\includegraphics[width=1\columnwidth]{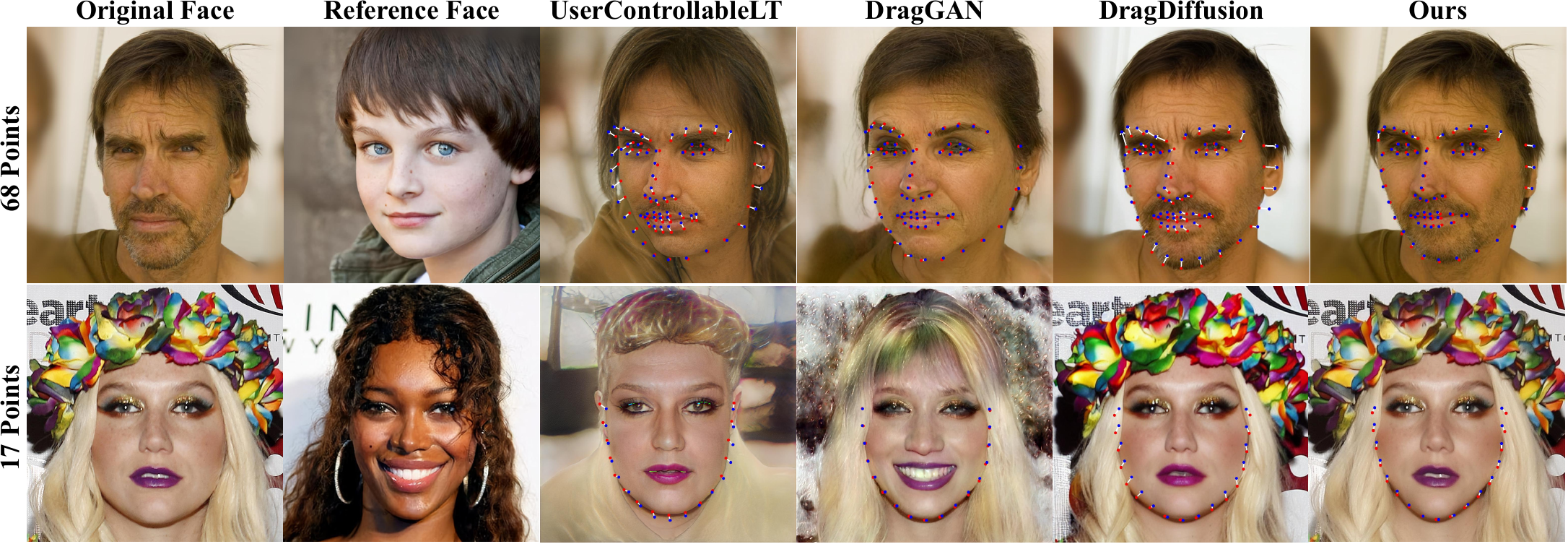}   
\end{minipage}
\centering
\vspace{-5pt}
\caption{Qualitative comparison between our DragonDiffusion and other methods in face manipulation. The current and target points are labeled with \textcolor{red}{red} and \textcolor{blue}{blue}. The white line indicates distance.}
\label{fig_cmp_face} 
\vspace{-10pt}
\end{figure*}

\subsection{Comparisons}
In this part, we compare our method with the recent UserControllableLT~\cite{user}, DragGAN~\cite{draggan}, and DragDiffusion~\cite{dragdiff} in the keypoint-based face manipulation.

\textbf{Time complexity}. We divide the time complexity of different methods into two parts, \textit{i.e.}, the preparing and inference stages. The preparing stage involves Diffusion/GAN inversion and model fine-tuning. The inference stage generates the editing result. The time complexity for each method is tested on one point dragging, with the image resolution being $512\times 512$. The experiment is conducted on an NVIDIA A100 GPU with Float32 precision. Tab.~\ref{tb_cp_face} presents that our method is relatively efficient in the preparing stage, requiring only 3.62s to prepare $\mathbf{z}_T$ and build a memory bank. The inference complexity is also acceptable for diffusion generation.

\textbf{Qualitative and quantitative evaluation}. Following DragGAN~\cite{draggan}, the comparison is conducted on the face keypoint manipulation with 17 and 68 points. The test set is randomly formed by 800 aligned faces from CelebA-HQ~\cite{celeba} training set. Note that we do not set fixed regions for all methods, due to the difficulty in manually providing a mask for each face. In addition to accuracy, we also compute the FID~\cite{fid} between face editing results and CelebA-HQ training set to represent the editing quality. The quantitative and qualitative comparison is presented in Tab.~\ref{tb_cp_face} and Fig.~\ref{fig_cmp_face}, respectively. We can find that although DragGAN can produce more accurate editing results, it has limitations in content consistency and robustness in areas outside faces (\textit{e.g.}, the headwear is distorted). 
The limitations of GAN models also result in DragGAN and UserControllableLT requiring face alignment before editing.
In comparison, our method has promising editing accuracy, and the generation prior from SD enables better robustness and generalization for different content. In this task, our method also has better performance than DragDiffusion. Moreover, the visual cross-attention design makes our method achieve attractive content consistency without extra model fine-tuning or modules. More results are shown in the \textbf{appendix}.

\begin{wrapfigure}{r}{0.45\textwidth}
    \centering
    \vspace{-10pt}
    \includegraphics[width=0.45\textwidth]{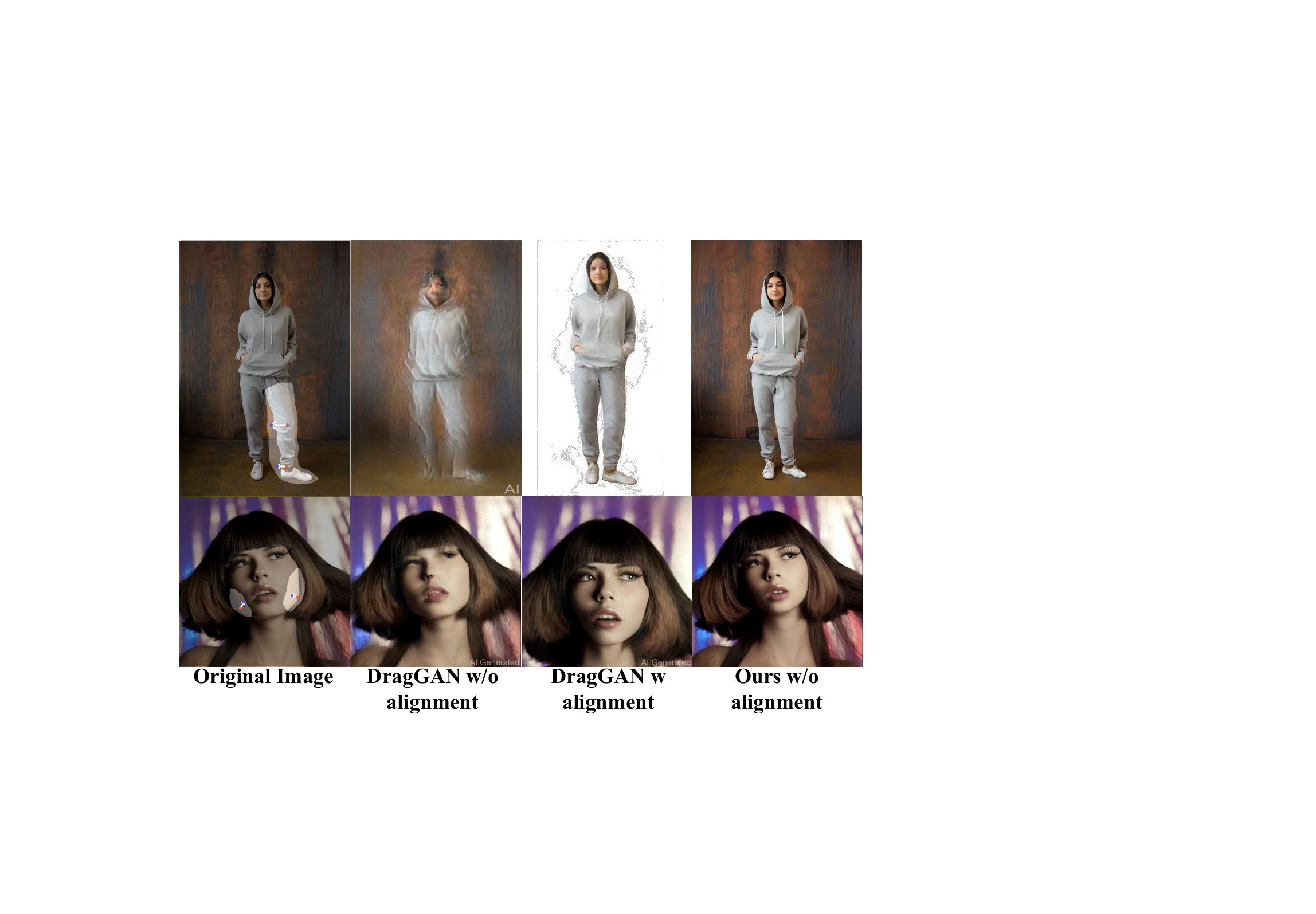}
    \vspace{-20pt}
    \caption{\small{Editing comparison between our DragonDiffusion and DragGAN~\cite{draggan} on the unaligned body and face.}}
    \label{fig_rebost} 
    \vspace{-30pt}
\end{wrapfigure}

\textbf{Further discussion on the generalization of DragGAN and our method}. 
Although DragGAN demonstrates powerful drag editing capabilities, its performance is significantly reduced in complex scenarios due to the limited capability of the GANs, as shown in Fig.~\ref{fig_rebost}. 
Specifically, we use the StyleGAN trained on the human bodies~\cite{stylegan-human} and FFHQ~\cite{ffhq} 
to perform body and face editing by DragGAN.
As can be seen, the editing quality of DragGAN is sensitive to whether the image is aligned. 
The alignment operation will filter out the background of the body or change the face pose, which usually does not meet our editing requirements. 
In comparison, our DragonDiffusion inherits good generalization of the pre-trained SD and can handle complex and unaligned scenarios effectively. The resolution of images processed by our method is also arbitrary, unlike the fixed size in GANs.

\subsection{Ablation Study}
\begin{figure*}[t]
\centering
\small 
\begin{minipage}[t]{\linewidth}
\centering
\includegraphics[width=1\columnwidth]{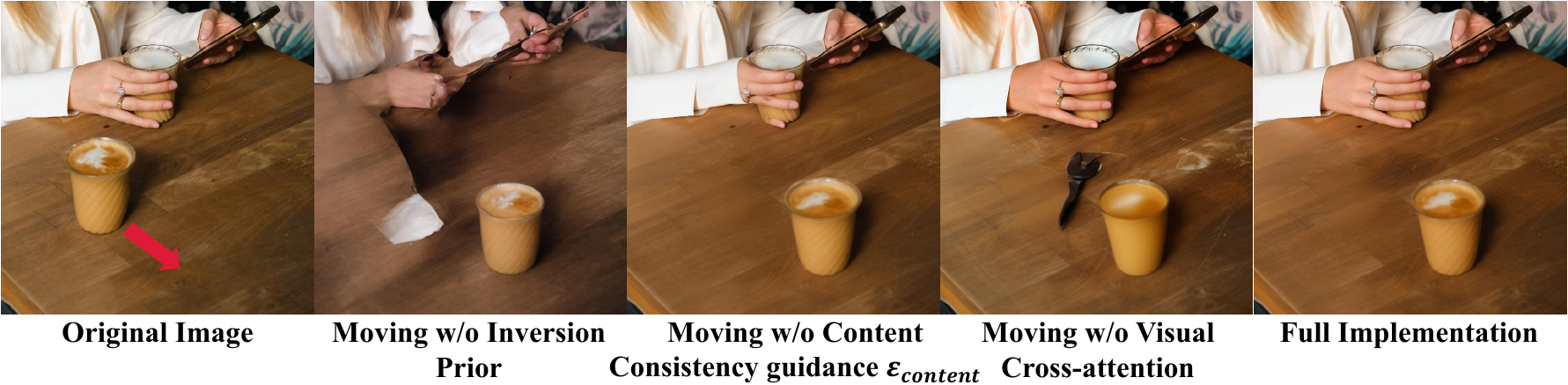}
\end{minipage}
\centering
\vspace{-15pt}
\caption{The ablation study of different components in our DragonDiffusion. The experiment is conducted on the task of object moving. }
\vspace{-10pt}
\label{fig_ab} 
\end{figure*}

In this part, we demonstrate the effectiveness of some components in our DragonDiffusion, as shown in Fig.~\ref{fig_ab}. We conduct the experiment on the object moving task. Specifically, \textbf{(1)} we verify the importance of the inversion prior by randomly initializing $\mathbf{z}_T$ instead of obtaining from DDIM inversion. As can be seen, the random $\mathbf{z}_T$ leads to a significant difference between the editing result and the original image. \textbf{(2)} We remove the content consistency guidance (\textit{i.e.}, $\mathcal{E}_{content}$) in Eq.~\ref{eq_loss}, which causes local distortion in the editing result, \textit{e.g.}, the finger is twisted. \textbf{(3)} We remove the visual cross-attention. It can be seen that visual cross-attention plays an important role in maintaining the consistency between the edited object and the original object. The last image shows the satisfactory editing performance of our method with full implementation. Therefore, these components work together on both edited and unedited content, forming the fine-grained image editing model DragonDiffusion, which does not require extra training or modules.

\section{Conclusion}
Despite the ability of existing large-scale text-to-image (T2I) diffusion models to generate high-quality images from detailed textual descriptions, they often lack the ability to precisely edit the generated or real images.
In this paper, we aim to develop a drag-style and general image editing scheme based on the strong correspondence of intermediate image features in the pre-trained diffusion model. To this end, we model image editing as the change of feature correspondence and design energy functions to transform the editing operations into gradient guidance.
Based on the gradient guidance strategy, we also propose multi-scale guidance to consider both semantic and geometric alignment. Moreover, a visual cross-attention is added based on a memory bank design, which can enhance the consistency between the original image and the editing result. Due to the reuse of intermediate information from the inversion process, this content consistency strategy almost has no additional cost. Extensive experiments demonstrate that our proposed DragonDiffusion can perform various image editing tasks, including object moving, resizing, appearance replacing, object pasting, and content dragging. At the same time, the complexity of our DragonDiffusion is acceptable, and it does not require extra model fine-tuning or additional modules.

\bibliography{iclr2024_conference}
\bibliographystyle{iclr2024_conference}

\clearpage

\appendix
\section{Appendix}
\subsection{Algorithm Pipeline of DragonDiffusion}
To facilitate the understanding of our DragonDiffusion, we present the entire algorithm pipeline in Algorithm~\ref{alg_1}. Note that the text condition $\mathbf{c}$ has a minimal impact on the final result in our method, and only a brief description of the image is needed. 

\begin{algorithm}[tbh]
\label{aug}
\caption{Proposed DragonDiffusion}
\textbf{Require}:

\hspace{0.01cm}text condition $\mathbf{c}$; UNet denoiser $\epsilon_{\boldsymbol{\theta}}$; pre-defined parameter $\bar{\alpha}_t$; image to be edited $\mathbf{x}_{0}$; the mask of $\mathbf{m}_{gen}$, $\mathbf{m}_{gud}$, and $\mathbf{m}_{share}$; the learning rate $\eta$; the number of gradient-guidance steps $n$.

\textbf{Initialization}:

    \hspace{0.01cm} (1) Compute latent $\mathbf{z}_{0}$ of the image to be edited:

    \hspace{0.5cm} $\mathbf{z}_{0} = Encoder(\mathbf{x}_{0})$

    \hspace{0.01cm} (2) Select an editing task $\text{Ts}\in [\text{`resizing}\&\text{moving'},\ \text{`dragging'},\ \text{`pasting'},\ \text{`replacing'}]$;

    \hspace{0.01cm} (3) Compute latent $\mathbf{z}_{0}^{ref}$ of the reference image $\mathbf{x}_{0}^{ref}$:
    
    \hspace{0.5cm}\textbf{if} $\text{Ts}\in [\text{`resizing}\&\text{moving'},\ \text{`dragging'}]$ \textbf{then} $\mathbf{z}_{0}^{ref}=\emptyset$ \textbf{else} $\mathbf{z}_{0}^{ref} = Encoder(\mathbf{x}_{0}^{ref})$ 
    
    \hspace{0.01cm} (4) Compute the inversion prior $\mathbf{z}_{T}^{gen}$ and build the memory bank:
    
    \hspace{0.5cm} $\mathbf{z}_{T}^{gen},\ \mathbf{Bank}=DDIMInversion(\mathbf{z}_{0},\ \mathbf{z}_{0}^{ref})$ 

 \For{$t=T,\ \ldots,\ 1$}{
 \eIf{$\text{Ts}\in [\text{`resizing}\&\text{moving'},\ \text{`dragging'}]$}{$\mathbf{K}_t^{gud},\ \mathbf{V}_t^{gud},\ \mathbf{z}_t^{gud} = \mathbf{Bank}[t]$;

 $\mathbf{K}_t^{ref},\ \mathbf{V}_t^{ref},\ \mathbf{z}_t^{ref}=\emptyset$;

 extract $\mathbf{F}_t^{gen}$ and $\mathbf{F}_t^{gud}$ from $\mathbf{z}_{t}^{gen}$ and $\mathbf{z}_{t}^{gud}$ by $\epsilon_{\boldsymbol{\theta}}$;

 $\mathbf{K}_t,\ \mathbf{V}_t = \mathbf{K}_t^{gud},\ \mathbf{V}_t^{gud}$;

 }{$\mathbf{K}_t^{gud},\ \mathbf{V}_t^{gud},\ \mathbf{z}_t^{gud},\ \mathbf{K}_t^{ref},\ \mathbf{V}_t^{ref},\ \mathbf{z}_t^{ref} = \mathbf{Bank}[t]$;

 extract $\mathbf{F}_t^{gen}$, $\mathbf{F}_t^{gud}$ and $\mathbf{F}_t^{ref}$ from $\mathbf{z}_{t}^{gen}$, $\mathbf{z}_{t}^{gud}$ and $\mathbf{z}_{t}^{ref}$ by $\epsilon_{\boldsymbol{\theta}}$;

 $\mathbf{K}_t,\ \mathbf{V}_t = \mathbf{K}_t^{gud}\circled{\textbf{c}}\mathbf{K}_t^{ref},\ \mathbf{V}_t^{gud}\circled{\textbf{c}}\mathbf{V}_t^{ref}$;

 }

 $\hat{\boldsymbol{\epsilon}}_t=\epsilon_{\boldsymbol{\theta}}(\mathbf{z}_{t}^{gen},\ \mathbf{K}_t,\ \mathbf{V}_t,\ t,\ \mathbf{c})$;

  \uIf{$T-t<n$}{$\mathcal{E}=w_{e}\cdot \mathcal{E}_{edit} + w_{c}\cdot \mathcal{E}_{content} + w_{o}\cdot \mathcal{E}_{opt}$;

 $\hat{\boldsymbol{\epsilon}_t} = \hat{\boldsymbol{\epsilon}}_t+\eta \cdot \nabla_{\mathbf{z}_{t}} \mathcal{E}$;}

 $\mathbf{z}_{t-1} = \sqrt{\bar{\alpha}_{t-1}}(\frac{\mathbf{z}_t-\sqrt{1-\bar{\alpha}_{t}}\hat{\boldsymbol{\epsilon}}_t}{\sqrt{\bar{\alpha}_{t}}}+\sqrt{1-\bar{\alpha}_{t-1}}\hat{\boldsymbol{\epsilon}}_t)$;
 }

\hspace{0.01cm} $\mathbf{x}_0 = Decoder(\mathbf{z}_0)$;

\textbf{Output:} $\mathbf{x}_0$
\label{alg_1}
\end{algorithm}
\vspace{-10pt}

\subsection{Efficiency of the Memory Bank Design}
\begin{figure*}[h]
\centering
\small 
\begin{minipage}[h]{\linewidth}
\centering
\includegraphics[width=1\columnwidth]{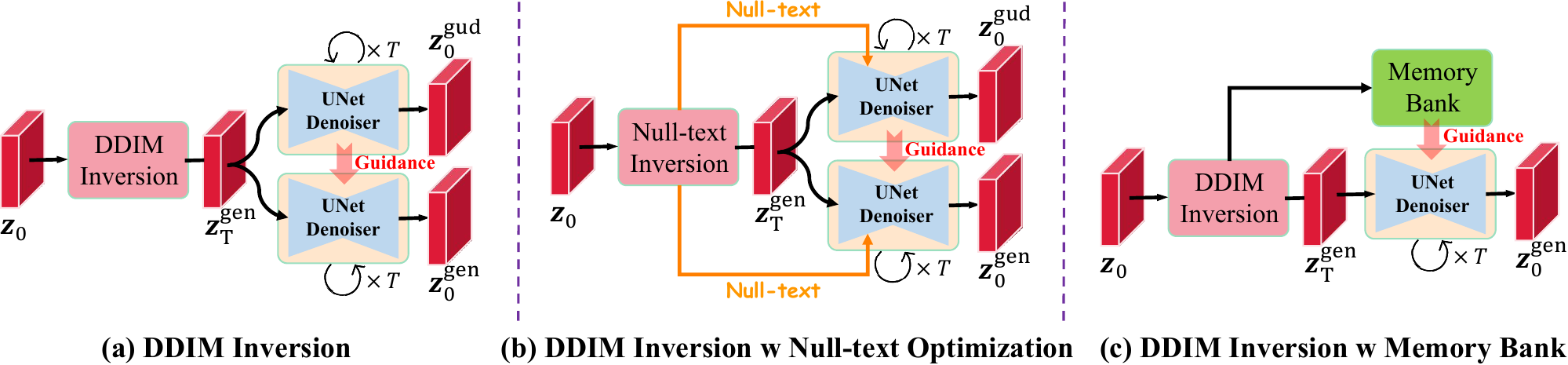}
\end{minipage}
\centering
\caption{Different strategies for generating inversion prior (\textit{i.e.}, $\mathbf{z}_T$) and guidance information (\textit{i.e.}, $\mathbf{K}_t^{gud}, \mathbf{V}_t^{gud}$). (a) DDIM inversion $+$ separate branch; (b) null-text inversion~\cite{null} $+$ separate branch; (c) our memory bank design.}
\label{fig_inv_model} 
\end{figure*}
\begin{figure*}[t]
\centering
\small 
\begin{minipage}[t]{\linewidth}
\centering
\includegraphics[width=1\columnwidth]{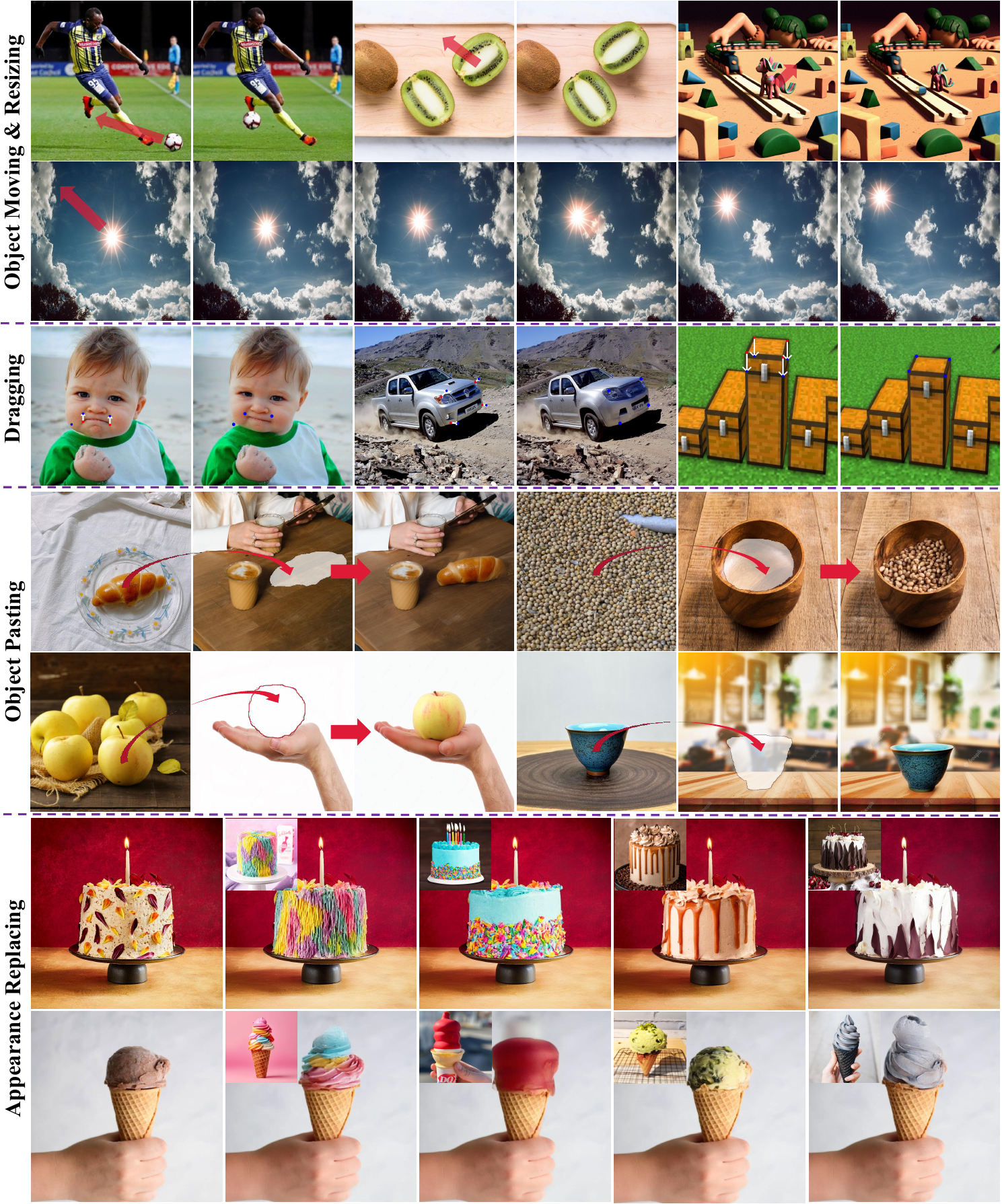}
\end{minipage}
\centering
\caption{More results of DragonDiffusion on different applications. }
\label{fig_supp} 
\end{figure*}

\begin{wrapfigure}{r}{0.45\textwidth}
    \centering
    \includegraphics[width=0.45\textwidth]{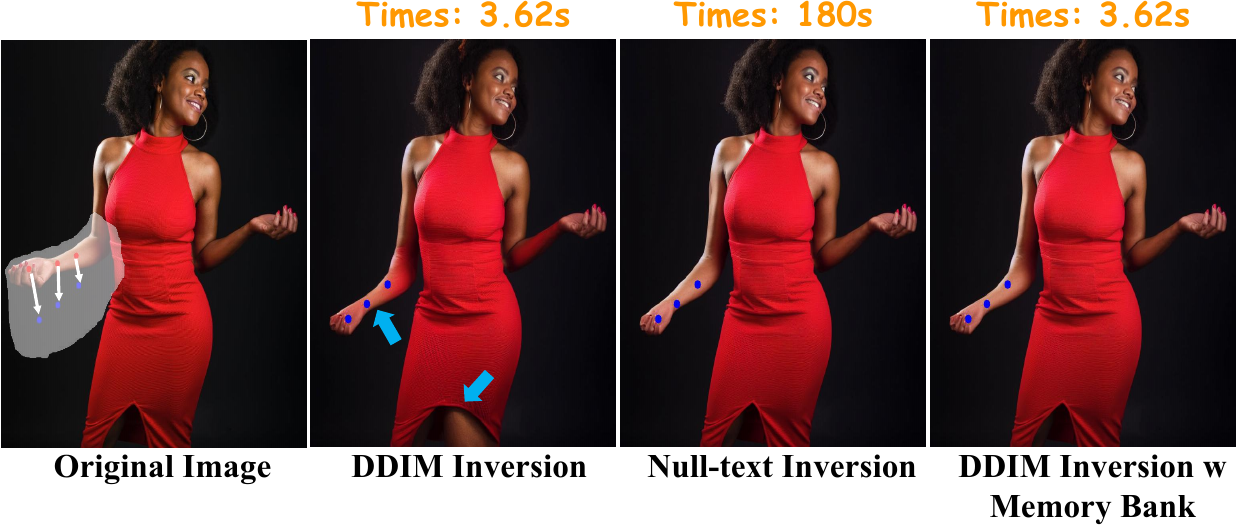}
    \vspace{-20pt}
    \caption{\small{The editing quality of different guidance strategies.}}
    \label{fig_inv_vis} 
    \vspace{-10pt}
\end{wrapfigure}
In this paper, we designed a memory bank to store intermediate information during the inversion process, which is used to provide guidance for image editing. To verify its effectiveness, we compared it with methods having the same function, as shown in Fig.~\ref{fig_inv_model}. Specifically, (a) guidance information is generated by a separate generation branch from $\mathbf{z}_{T}$; (b) null-text optimization is added based on method (a); (c) using our designed memory bank strategy. The editing quality of different methods is presented in Fig.~\ref{fig_inv_vis}. It can be seen that extracting guidance information from $\mathbf{z}_{T}$ using a separate branch can lead to deviations. This is due to the approximation bias in DDIM inversion. Although incorporating null-text optimization can yield more accurate results, it comes with higher time complexity. Our method tactfully utilizes a memory bank to store intermediate information during the inversion process, achieving accurate results while maintaining a time complexity of only $3.62$ seconds.

\subsection{More Results of DragonDiffusion on Different Applications}
In this part, we present more visual results of our DragonDiffusion in different applications, as shown in Fig.~\ref{fig_supp}. The first and second rows show the visualization of our object moving performance. It can be seen that our method has attractive object moving performance and good content consistency even in complex scenarios. The continuous moving editing presents attractive editing stability. The third row demonstrates that our method can perform natural point-drag editing of image content in different scenarios with several points. The fourth and fifth rows show the performance of our method in cross-image object pasting tasks. It can be seen that our method can fine-tune an object in one image and then naturally paste it onto another image. The last two rows demonstrate the performance of our method in object appearance replacing. It can be seen that our DragonDiffusion not only has good editing quality on small objects (\textit{e.g.}, ice-cream) but also performs well in replacing the appearance of large objects (\textit{e.g.}, cakes). Therefore, without any training and additional modules, our DragonDiffusion performs well in various image editing tasks. 

\subsection{More Qualitative Comparisons between Our DragonDiffusion and other Methods on Content Dragging}
\begin{figure*}[h]
\centering
\small 
\begin{minipage}[h]{\linewidth}
\centering
\includegraphics[width=1\columnwidth]{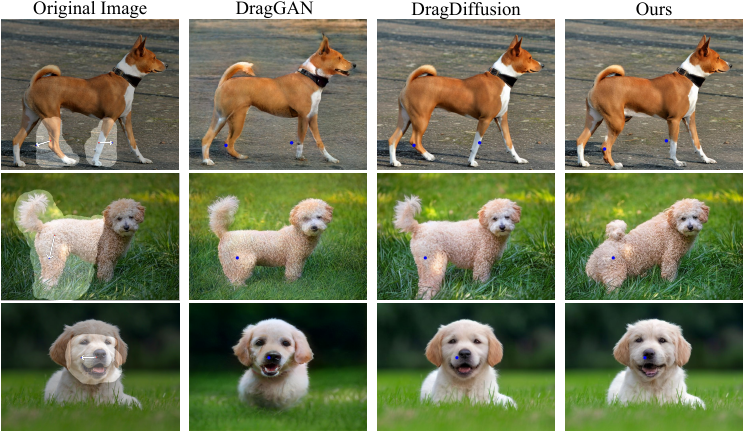}
\end{minipage}
\centering
\caption{
More qualitative comparison between our DragonDiffusion and other methods on the dog dragging. 
It can be seen that DragGAN~\cite{draggan} is limited in generation quality and content consistency due to the capabilities of GAN models. DragDiffusion~\cite{dragdiff} experiences an accuracy decline when dealing with larger editing drags, such as changing the posture of the dog's body. In comparison, our method has promising performance in these aspects.
}
\label{fig_ap-dog} 
\end{figure*}

\begin{figure*}[h]
\centering
\small 
\begin{minipage}[h]{\linewidth}
\centering
\includegraphics[width=1\columnwidth]{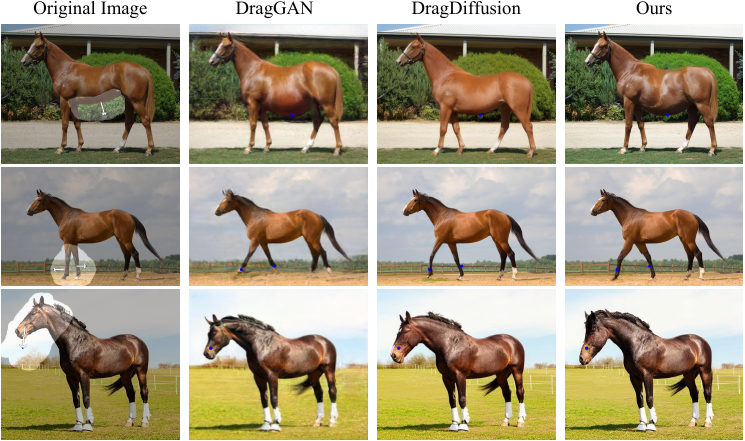}
\end{minipage}
\centering
\caption{More qualitative comparison between our DragonDiffusion and other methods on the horse dragging.}
\label{fig_ap-horse} 
\end{figure*}

\begin{figure*}[h]
\centering
\small 
\begin{minipage}[h]{\linewidth}
\centering
\includegraphics[width=1\columnwidth]{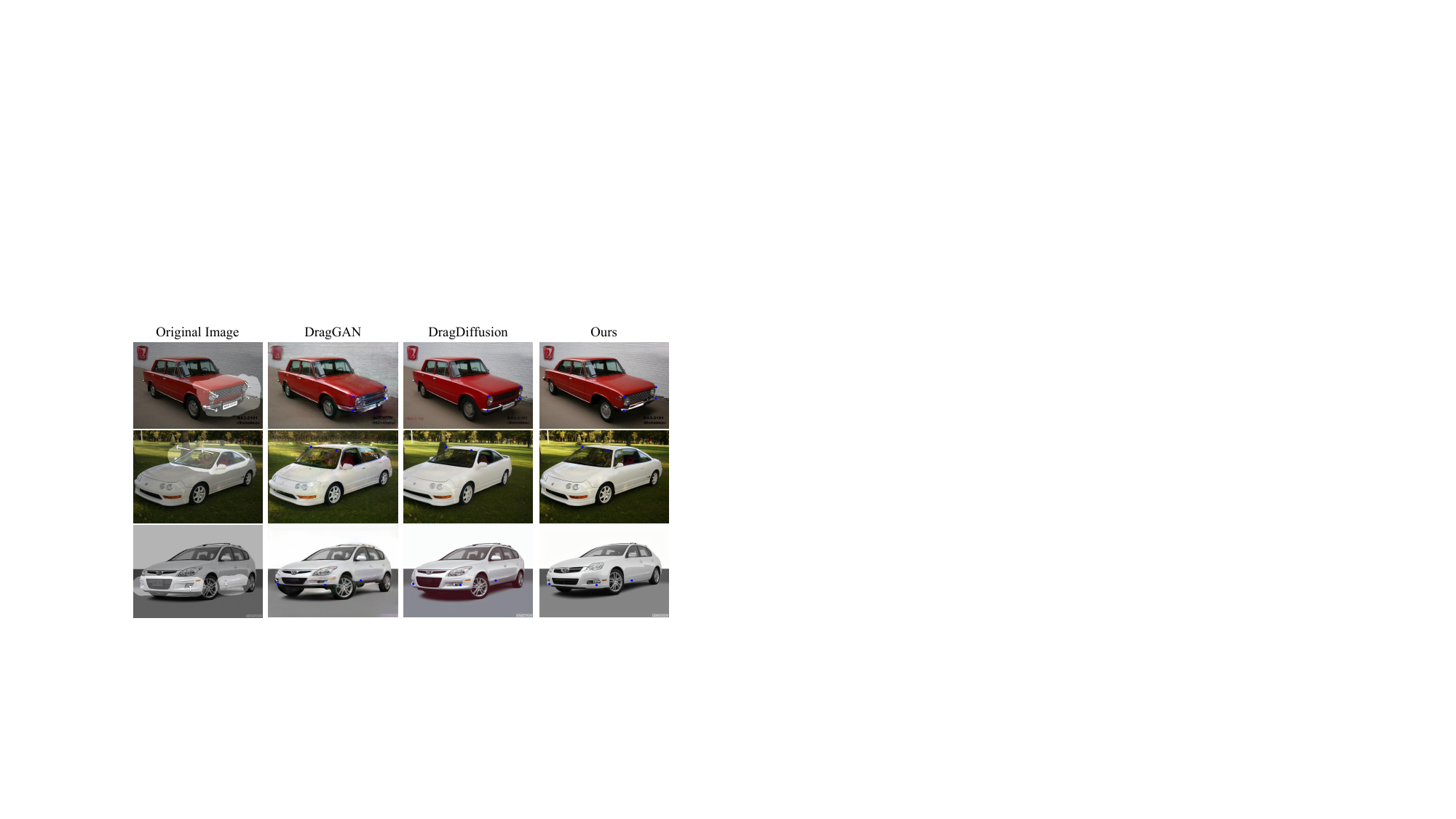}
\end{minipage}
\centering
\caption{More qualitative comparison between our DragonDiffusion and other methods on the car dragging.}
\label{fig_ap-car} 
\end{figure*}

\begin{figure*}[h]
\centering
\small 
\begin{minipage}[h]{\linewidth}
\centering
\includegraphics[width=1\columnwidth]{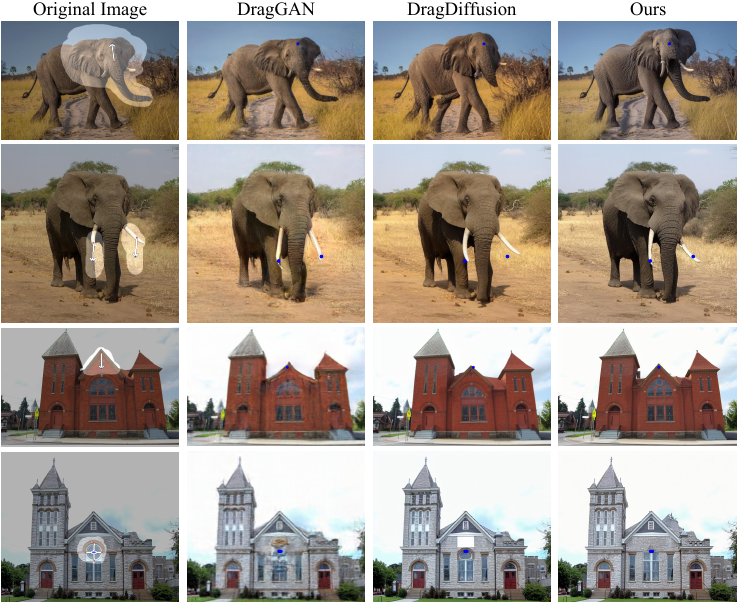}
\end{minipage}
\centering
\caption{More qualitative comparison between our DragonDiffusion and other methods on the church and elephant dragging.}
\label{fig_ap-church} 
\end{figure*}

\begin{figure*}[h]
\centering
\small 
\begin{minipage}[h]{\linewidth}
\centering
\includegraphics[width=1\columnwidth]{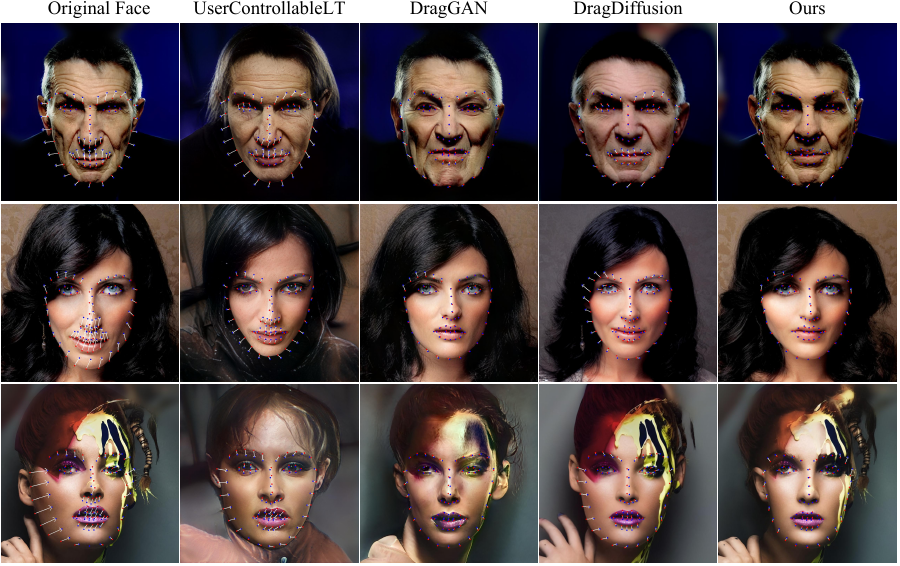}
\end{minipage}
\centering
\caption{More qualitative comparison between our DragonDiffusion and other methods on the face dragging.}
\label{fig_ap-face} 
\end{figure*}

In this part, we demonstrate more qualitative comparisons between our DragonDiffusion and other methods
on more categories. Fig.~\ref{fig_ap-dog} shows the comparison of drag editing on dogs. Fig.~\ref{fig_ap-horse} shows the comparison of drag editing on horses. Fig.~\ref{fig_ap-car} shows the comparison of drag editing on cars. Fig.~\ref{fig_ap-church} shows the comparison of drag editing on churches and elephants. Fig.~\ref{fig_ap-face} shows the comparison of drag editing on face manipulation. In these comparisons, DragGAN~\cite{draggan} requires switching between different models for different categories. Our method and DragDiffusion~\cite{dragdiff} benefit from the powerful generalization capabilities of SD~\cite{ldm}, enabling a single model to address image editing across different categories. These visualization results show that our method can produce better consistency with original images. At the same time, our method well balances the editing accuracy and generation quality.

\subsection{User Study}
\begin{figure*}[h]
\centering
\small 
\begin{minipage}[h]{\linewidth}
\centering
\includegraphics[width=1\columnwidth]{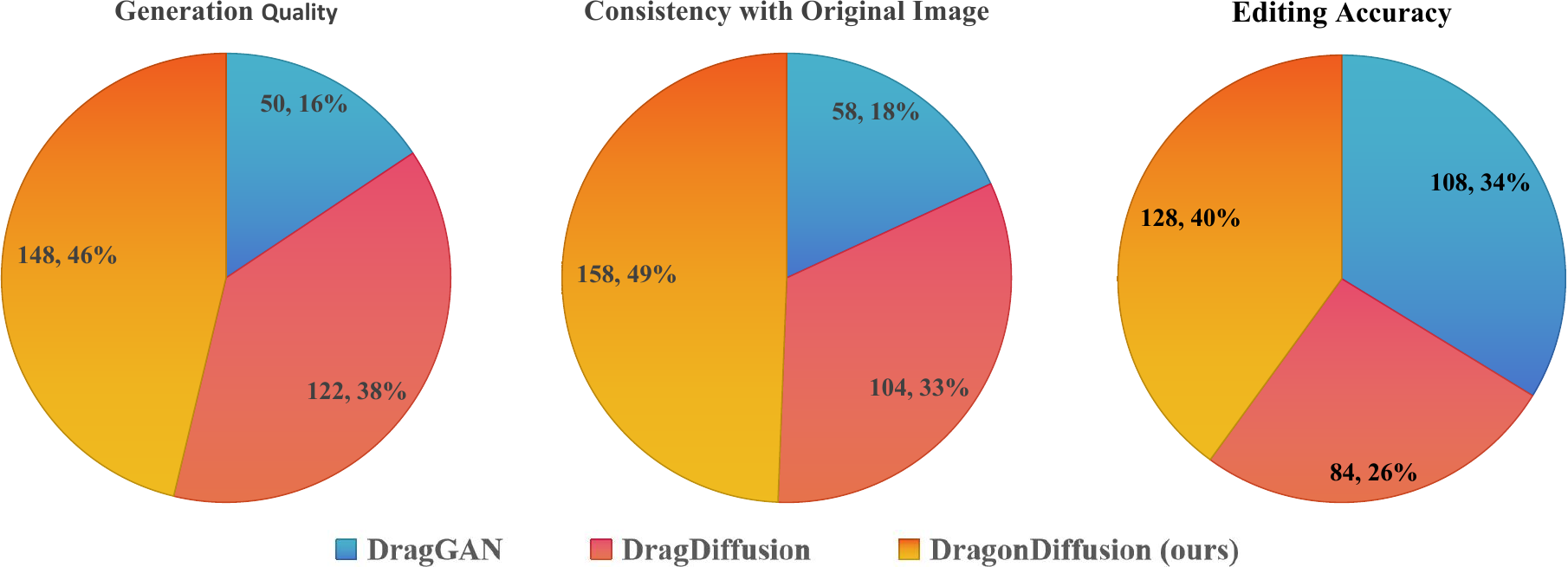}
\end{minipage}
\centering
\caption{User study of DragGAN~\cite{draggan}, DragDiffusion~\cite{dragdiff}, and our DragonDiffusion. The experiment is conducted on various categories including dog, horse, car, elephant, church and face.}
\label{fig_user} 
\end{figure*}

To further compare with DragGAN~\cite{draggan} and DragDiffusion~\cite{dragdiff}, we design a user study, which includes three evaluation aspects: generation quality, editing accuracy, and content consistency. The test samples involve various categories including dog, horse, car, elephant, church, and face. We allow 20 volunteers to choose the best-performing method in each of the 16 groups of images and then compile the votes in Fig.~\ref{fig_user}. As can be seen, our method has better subjective performance in these three aspects.

\subsection{Demo Video}
A demo video is attached to the supplementary materials.

\end{document}